\documentclass[runningheads]{llncs}
\usepackage{graphicx}
\usepackage{tikz}
\usepackage{comment} 
\usepackage{amsmath,amssymb} % define this before the line numbering.
\usepackage{color}
\usepackage{epsfig}
\usepackage{graphicx}
\usepackage{makecell}
\usepackage{multirow}
\usepackage{float}
\newcommand{\tablestyle}[2]{\setlength{\tabcolsep}{#1}\renewcommand{\arraystretch}{#2}\centering\footnotesize}
\newcolumntype{x}[1]{>{\centering\arraybackslash}p{#1pt}}
\usepackage{verbatim}
\usepackage{enumitem}
\usepackage{booktabs}
\usepackage{tabulary,multirow,overpic}
\usepackage[caption=false]{subfig}
\usepackage{hyperref}
\definecolor{deeppink}{rgb}{1.0, 0.08, 0.58}
\hypersetup{
    colorlinks=true,
    urlcolor=deeppink,
}
\usepackage{xcolor}
\usepackage{authblk}

\begin{document}
% \linenumbers
\pagestyle{headings}
\mainmatter
\def\ECCVSubNumber{100}  % Insert your submission number here

\title{Forecasting Human-Object Interaction: \\
Joint Prediction of Motor Attention \\and Actions in First Person Video} % Replace with your title

\begin{comment}
\titlerunning{ECCV-20 submission ID \ECCVSubNumber} 
\authorrunning{ECCV-20 submission ID \ECCVSubNumber} 
\author{Anonymous ECCV submission}
\institute{Paper ID \ECCVSubNumber}
\end{comment}
%******************

%\begin{comment}
\titlerunning{Forecasting Human-Object Interaction}
\author{Miao Liu\inst{1}\and
Siyu Tang\inst{3}\and
Yin Li\inst{2}\and
James M. Rehg\inst{1}}
\authorrunning{M. Liu et al.}
\institute{Georgia Institute of Technology, Atlanta, United States\\ \and
University of Wisconsin-Madison, Madison, United States\\\and
ETH Z\"urich, Switzerland
}

%\end{comment}
%******************
\maketitle

\begin{abstract}
We address the challenging task of anticipating human-object interaction in first person videos. Most existing methods either ignore how the camera wearer interacts with objects, or simply considers body motion as a separate modality. In contrast, we observe that the intentional hand movement reveals critical information about the future activity. Motivated by this observation, we adopt intentional hand movement as a feature representation, and propose a novel deep network that jointly models and predicts the egocentric hand motion, interaction hotspots and future action. Specifically, we consider the future hand motion as the motor attention, and model this attention using probabilistic variables in our deep model. The predicted motor attention is further used to select the discriminative spatial-temporal visual features for predicting actions and interaction hotspots. We present extensive experiments demonstrating the benefit of the proposed joint model. Importantly, our model produces new state-of-the-art results for action anticipation on both EGTEA Gaze+ and the EPIC-Kitchens datasets. Our project page is available at \url{https://aptx4869lm.github.io/ForecastingHOI/}
\keywords{First Person Vision, Action Anticipation, Motor Attention}
\end{abstract}

\section{Introduction}

\begin{figure}[t]
\centering

\includegraphics[width=0.85\linewidth]{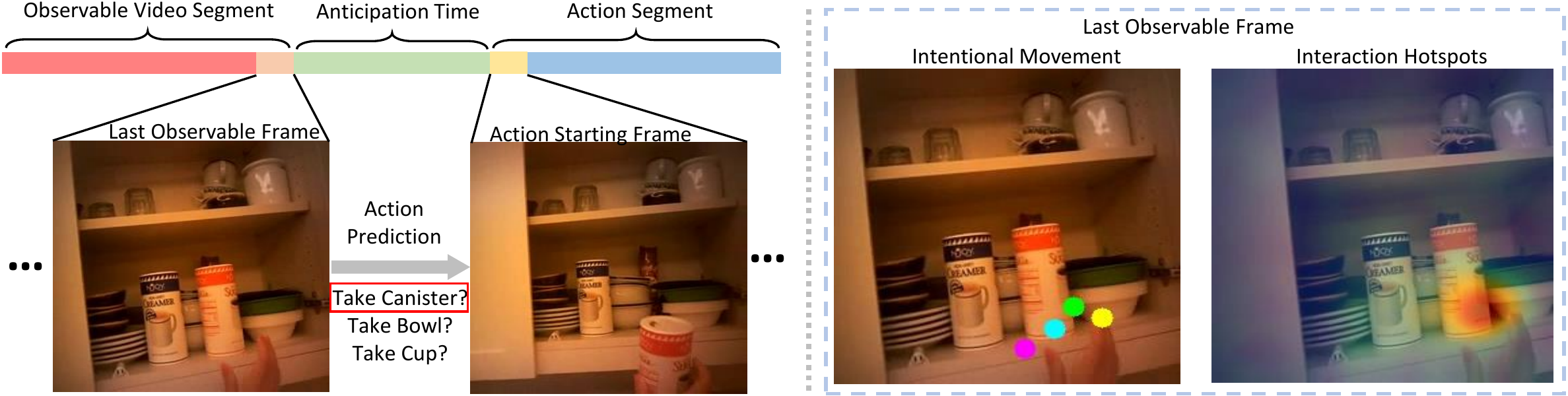}
\caption{{\it What is the most likely future interaction?} Our model takes advantage of the connection between motor attention and visual perception. In addition to future action label, our model also predicts the interaction hotspots on the last observable frame and hand trajectory (in the order of {\color{yellow} yellow} , {\color{green} green}, {\color{cyan} cyan}, and {\color{magenta} magenta}) between the last observable time step to action starting point. Visualizations of hand trajectory are projected to the last observable frame (best viewed in color).} 
\label{fig:teaser}

\end{figure}

The human ability of ``looking into the near future'' remains a key challenge for computer vision. Consider the example in Fig.\ \ref{fig:teaser}, given a video shortly before the start of an action, we can easily predict what will happen next, e.g., the person will take the canister of salt. Even without seeing any future frames, we can vividly imagine how the person will perform the action, e.g., the trajectory of the hand when reaching for the canister or the location on the canister that will be grasped.

There is convincing evidence that our remarkable ability to forecast other individuals' actions depends critically  upon our perception and interpretation of their body motion. The investigation of this anticipatory mechanism dates back to $19$th century, when William James argued that future expectations are intrinsically related to purposive body movements~\cite{james1890principles}. 
Additional evidence for a link between perceiving and performing actions was provided by the discovery of mirror neurons~\cite{di1992understanding,hari1998activation}. 
The observation of others' actions activates our motor cortex, the same brain regions that are in charge of the planning and control of intentional body motion. 
This activation can happen even before the onset of the action and is highly correlated with the anticipation accuracy~\cite{aglioti2008action}. 
A compelling explanation from~\cite{rushworth2003left} suggests that {\it motor attention}, i.e., the active prediction of meaningful future body movements, serves as a key representation for anticipation. A goal of this work is to develop a computational model for motor attention that can enable more accurate action prediction. 

Despite these relevant findings in cognitive neuroscience, the role of intentional body motion in action anticipation is largely ignored by the existing literature~\cite{vondrick2016anticipating,Felsen_2017_ICCV,gao2017red,kataoka2016recognition,furnari2017next,furnari2019rulstm,Miech_2019_CVPR_Workshops,Ke_2019_CVPR}. In this work, we focus on the problem of forecasting human-object interactions in First Person Vision (FPV). Interactions consist of a single verb and one or more nouns, with ``take bowl'' as an example. FPV videos capture complex hand movements during a rich set of interactions, thus providing a powerful vehicle for studying the connection between motor attention and future representation. Several previous works have investigated the problems of FPV activity anticipation~\cite{furnari2017next,furnari2019rulstm} and body movement prediction~\cite{aksan2019structured,gui2018adversarial,fragkiadaki2015recurrent,walker2017pose}. We believe we are the first to utilize a motor attention model for FPV action anticipation.

To this end, we propose a novel deep model that predicts ``motor attention''---the future trajectory of the hands, as an anticipatory representation of actions. Based on motor attention, our model further localizes the future contact region of the interaction, i.e., interaction hotspots~\cite{nagarajan2018grounded} and recognizes the type of future interactions. Importantly, we characterize motor attention and interaction hotspots as probabilistic variables modeled by stochastic units in a deep network. These units naturally deal with the uncertainty of future hand motion and contact region during interaction, and produce attention maps that highlight discriminative spatial-temporal features for action anticipation. 

% we still need to talk about the results
During inference, our model takes video clips shortly before the interaction as inputs, and jointly predicts motor attention, interaction hotspots, and action labels. During training, our model assumes that these outputs are available as supervisory signals. To evaluate our model, we report results on two major FPV benchmarks: EGTEA Gaze+ and EPIC-Kitchens. Our approach outperforms prior state-of-the-art methods by a significant margin. In addition, we conduct extensive ablation studies to verify the design of our model and evaluate our model for motor attention prediction and interaction hotspots estimation. Our model demonstrates strong results for both tasks. We believe our model provides a solid step towards the challenge of FPV visual anticipation.

% Our main contributions are summarized as follows: (1)We present the first joint model that predicts motor attention, interaction hotspots and actions in FPV. (2)We proposed a novel probabilistic deep model that utilizes variational learning and approximate inference for joint modeling and stochastic units to account for the uncertainty of future representation. (3)We demonstrate that modeling motor attention is important for visual anticipation in FPV, leading to state-of-the-art results on major benchmarks.

\section{Related Works}

There has recently been substantial interest in learning to forecast future events in videos. The most relevant works to ours are those investigations on FPV action anticipation. Our work is also related to previous studies on third person action anticipation, other visual prediction tasks, and visual affordance.

\noindent \textbf{FPV Action Anticipation}.
Action anticipation aims at predicting an action before it happens. We refer the readers to a recent survey~\cite{kong2018human} for a distinction between action recognition and anticipation. FPV action recognition has been studied extensively~\cite{ryoo2015pooled,poleg2016compact,fathi2011understanding,Zhou_2016_CVPR,ma2016going,li2015delving,Li_2018_ECCV,pirsiavash2012detecting}, while fewer works have targeted egocentric action anticipation. Shen et al.\ \cite{shen2018egocentric} investigated how different egocentric modalities affect the action anticipation performance. Soran et al.\ \cite{soran2015generating} adopted Hidden Markov Model to compute the transition probability among sequences of actions. A similar idea was explored in~\cite{Miech_2019_CVPR_Workshops}. Furnari et al.\ \cite{furnari2017next} considered the task of predicting the next-active objects. Their recent work~\cite{furnari2019rulstm} proposed to factorize the anticipation model into a ``Rolling" LSTM that summarizes the past activity and an ``Unrolling" LSTM that makes hypotheses of the future activity. Ke et al.\ \cite{Ke_2019_CVPR} proposed a time-conditioned skip connection operation to extract relevant information for action anticipation. In contrast to our proposed method, these prior works did not exploit the connection between human motor attention and visual perception, and did not explicitly model the contact region during human-object interaction. 

\noindent \textbf{Third Person Action Anticipation}.
Several previous efforts seek to address the task of action anticipation in third person vision. Kris et al.\ \cite{kitani2012activity} combined semantic scene labeling with a Markov decision process to forecast the behavior and  trajectory of a subject. Vondrick et al.\ \cite{vondrick2016anticipating} proposed to predict the future video representation from large scale unlabeled video data. Gao et al.\ \cite{gao2017red} proposed a Reinforced Encoder-Decoder network to create a summary representation of past frames and produce a hypothesis of future action. Kataoka et al.\ \cite{kataoka2016recognition} introduced a subtle motion descriptor to identify the difference between an on-going action and a transitional action, and thereby facilitate future anticipation. Our work shares the same goal of future forecasting, but we focus on leveraging abundant visual cues from egocentric videos for action anticipation.

\noindent \textbf{Other Prediction Tasks}. 
Anticipation has been studied under other vision tasks. In particular, human body motion prediction has been extensively studied~\cite{pavlovic2001learning,urtasun2008topologically,wang2007gaussian,gui2018adversarial,walker2017pose,fragkiadaki2015recurrent}, including recent work in the setting of FPV. Rhinehart et al.\ \cite{Rhinehart_2017_ICCV} proposed an online learning algorithm to forecast the first-person trajectory. Park et al.\ \cite{soo2016egocentric} proposed a deep network to infer possible human trajectories from egocentric stereo images. Wei et al.\ \cite{wei2017inferring} utilized a probabilistic model to infer 3D human attention and intention. Tagi et al.\ \cite{yagi2018future} addressed a novel task of predicting the future locations of an observed subject in egocentric videos. Ryoo et al.\ \cite{ryoo2015robot} proposed a novel method to summarize pre-activity observations for robot-centric activity prediction. However, none of these previous work considered modeling body movement for action anticipation.

\noindent \textbf{Visual Affordance}. 
The problem of predicting visual affordances has attracted growing interest in computer vision. Affordance can be helpful for scene understanding~\cite{grabner2011makes,delaitre2012scene,wang2017binge}, human-object interaction recognition~\cite{thermos2017deep}, and action analysis~\cite{rhinehart2016learning,koppula2015anticipating}. Several recent works have focused on estimating visual affordances that are grounded on human object interaction. Chen et al.\ \cite{chen2018subjects} proposed to estimate likely object interaction regions by learning the connection between subject and object. Fang et al.\ \cite{fang2018demo2vec} proposed to estimate interaction regions by learning from demonstration videos. However, none of these previous works considered future prediction. More recently, Tushar et al.\ \cite{nagarajan2018grounded} introduced an unsupervised learning method that uses the backward attention map to approximate the interaction hotspots grounded on a future action. However, their method did not model the presence of objects and thus can not be used to anticipate human-object interactions. However, we compare to their results for interaction hotspot estimation in our experiments.

\section{Method}
%\miao{Why don't we just use the same mathematical expression as EPIC-Kitchens for problem define? This current form implies $\Delta \tau_a$ is derived from $\tau_a$. In fact, $\Delta \tau_a$ is fixed and $\tau_a$ is derived from $\Delta \tau_a$ .}
%\yin{The definition from EPIC-Kitchens is somewhat vague. Prediction is fundamentally a ``look-ahead'' problem. If you start with a pre-defined action segment, and define prediction as using a previous video clip to predict current action label. It seems less convincing to me.}

We consider the setting of action anticipation from~\cite{Damen2018EPICKITCHENS}. Denote an input video segment as $x:[\tau_a - \Delta \tau_o, \tau_a]$. $x$ starts at $\tau_a - \Delta \tau_o$ and ends at $\tau_a$ with duration $\Delta \tau_o>0$ as the ``observation time''. Our goal is to predict the label $y$ of an immediate future interaction starting at $\tau_s = \tau_a + \Delta \tau_a$, where $\Delta \tau_a>0$ is a fixed interval known as the ``anticipation time.''  Moreover, we seek to estimate future hand trajectories $\mathcal{M}$ within $[\tau_a, \tau_s]$ (projected back to the last observable frame at $\tau_a$), and to localize interaction hotspots $\mathcal{A}$ at $\tau_a$ (the last observable frame). Fig.\ \ref{fig:teaser} illustrates our setting.
%Similar to previous work~\cite{Damen2018EPICKITCHENS,furnari2019rulstm}, we focus on short-term anticipation with $\Delta \tau_a \le 1$ second.

\begin{figure*}[t]
\centering
\includegraphics[width=0.85\linewidth]{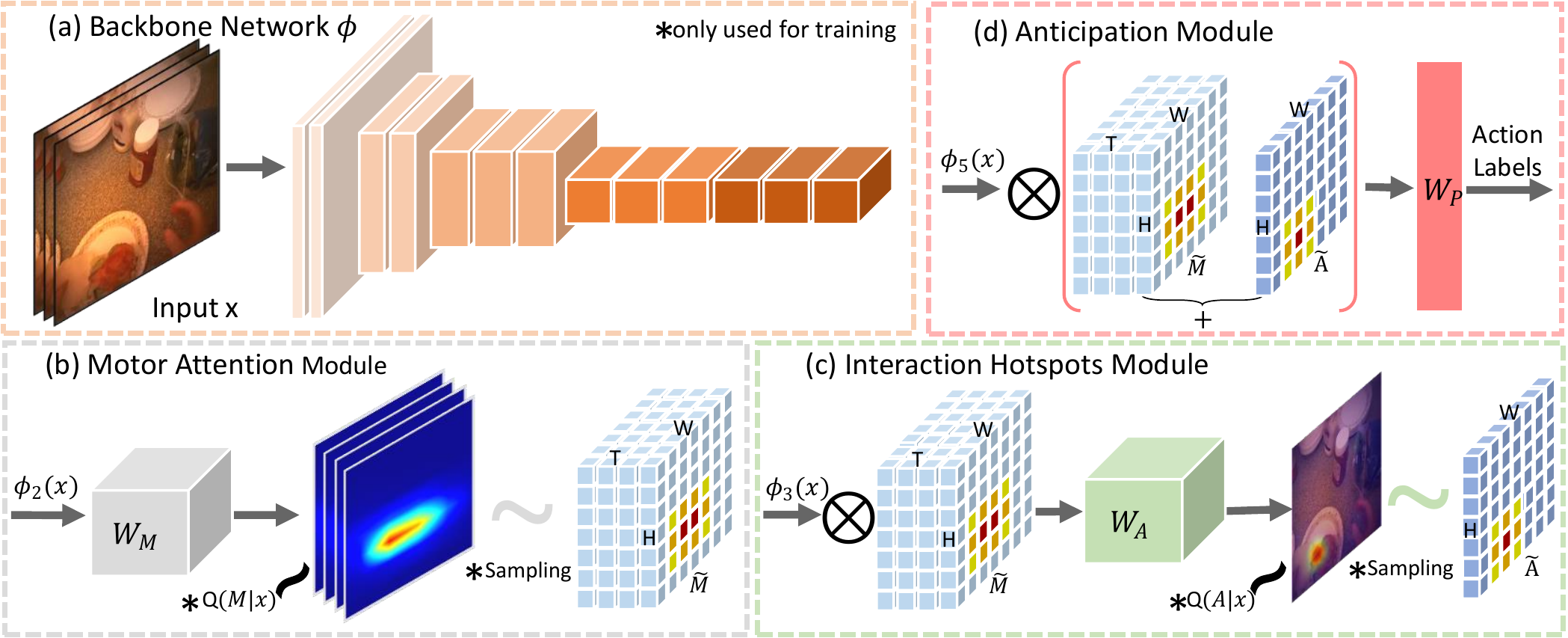}
\caption{Overview of our model. A 3D convolutional network $\phi(x)$ is used as our backbone network, with features from its $i^{th}$ convolution block as $\phi_i(x)$ (a). A motor attention module (b) makes use of stochastic units to generate sampled future hand trajectories $\tilde{\mathcal{M}}$ used to guide interaction hotspots estimation in module (c). Module (c) further generates sampled interaction hotspots $\tilde{\mathcal{A}}$ with a similar stochastic units as in module (b). Both $\tilde{\mathcal{M}}$ and $\tilde{\mathcal{A}}$ are used to guide action anticipation in anticipation module (d). During testing, our model takes only video clips as inputs, and predicts motor attention, interaction hotspots, and action labels. Note that $\otimes$ represents element-wise multiplication for weighted pooling.}
\label{fig:overview}
\end{figure*}

To summarize, our model seeks to anticipate the future action $y$ by jointly predicting the future hand trajectory $\mathcal{M}$ and interaction hotspots $\mathcal{A}$ at the last observable frame. Predicting the future is fundamentally ambiguous, since the observation of future interaction only represents one of the many possibilities characterized by an underlying distribution. Our key idea is thus to model motor attention and interaction hotspots as probabilistic variables in order to account for their uncertainty. We present an overview of our model in Fig.\ \ref{fig:overview}. 

Specifically, we make use of a 3D backbone network $\phi(x)$ for video representation learning. Following the approach in~\cite{simonyan2014very,he2016deep}, we utilize 5 convolutional blocks, and denote the features from the $i^{th}$ convolution block as $\phi_i(x)$. Based on $\phi(x)$, our motor attention module (b) predicts future hand trajectories as motor attention $\mathcal{M}$ and uses stochastic units to sample from $\mathcal{M}$. The sampled motor attention $\tilde{\mathcal{M}}$ is an indicator of important spatial-temporal features for interaction hotspot estimation. Our interaction hotspot module (c) further produces an interaction hotspot distribution $\mathcal{A}$ and its sample $\tilde{\mathcal{A}}$. Finally, our anticipation module (d) makes use of both $\tilde{\mathcal{M}}$ and $\tilde{\mathcal{A}}$ to aggregate network features, and predicts the future interaction $y$.

% Assume an action segment starts at time step $\tau_s$. Our goal is to predict the activity labels by observing a $\tau_o$ seconds video clip that precedes the action by $\tau_a$ seconds. Formally, we denote $\tau_a$ as the ``anticipation time'' and $\tau_o$ as the ``observation time''. Given the observable video segment $x:\{\tau_s-(\tau_a+\tau_o),\tau_s-\tau_a\}$, we seek to predict the action label $y$ that begins at $\tau_s$. In addition, our model also outputs the interaction hotspots $\mathcal{A}$ at time step $\tau_s-\tau_a$ (the last observable frame), and motor attention $\mathcal{M}$ from time step $\tau_s-\tau_a$ to time step $\tau_s$. We refer readers to Fig.\ \ref{fig:teaser} for a visual illustration of our problem setting.

\subsection{Joint Modeling of Human-Object Interaction}
Formally, we consider motor attention $\mathcal{M}$ and interaction hotspots $\mathcal{A}$ as probabilistic variables, and model the conditional probability of the future action label $y$ given the input video $x$ as a latent variable model, where
\begin{equation}
\small
p(y|x) = \int_{\mathcal{M}}  \int_{\mathcal{A}} p(y|\mathcal{A},\mathcal{M}, x) p(\mathcal{A}|\mathcal{M},x) p(\mathcal{M}|x) \ d\mathcal{A} \ d\mathcal{M}, 
\end{equation}
$p(\mathcal{M}|x)$ first estimates motor attention from video input $x$. $\mathcal{M}$ is further used to estimate interaction hotspots $A$ ($p(\mathcal{A}|\mathcal{M},x)$). Given $x$, $\mathcal{M}$ and $\mathcal{A}$, the action label $y$ is determined by $p(y|\mathcal{A},\mathcal{M}, x)$. Our model thus consists of three main components. 

\noindent \textbf{Motor Attention Module} tackles $p(\mathcal{M}|x)$. Given the network features $\phi_2(x)$, our model uses a function $F_M$ to predict motor attention $\mathcal{M}$. $\mathcal{M}$ is represented as a 3D tensor of size $T_m \times H_m \times W_m$. Moreover, $\mathcal{M}$ is normalized within each temporal slice, i.e., $\sum_{w,h} \mathcal{M}(t,w,h)=1$. 
%$\mathcal{M}$ is thus a sequence of 2D attention maps $\mathcal{M}(t)$ defined over $T_m$ steps. 

\noindent \textbf{Interaction Hotspots Module} targets at $p(\mathcal{A}|\mathcal{M},x)$. Our model uses a function $F_A$ to estimate the interaction hotspots $\mathcal{A}$ based on the network feature $\phi_3(x)$ and sampled motor attention $\tilde{\mathcal{M}}$. $\mathcal{A}$ is represented as a 2D attention map of size $H_a \times W_a$. A further normalization constrained that $\sum_{w,h} \mathcal{A}(w,h)=1$. 

\noindent \textbf{Anticipation Module} makes use of the predicted motor attention and interaction hotspots for action anticipation. Specifically, sampled motor attention $\tilde{\mathcal{M}}$ and sampled interaction hotspots $\tilde{\mathcal{A}}$ are used to aggregate feature $\phi_5(x)$ via weighted pooling. An action anticipation function $F_P$ further maps the aggregated features to future action label $y$.

\subsection{Motor Attention Module}
\noindent \textbf{Motor Attention Generation}. The motor attention prediction function $F_M$ is composed of a linear function with parameter $W_M$ on top of network features $\phi_2(x)$. The linear function is realized by a 3D convolution and a softmax function is used to normalized the attention map. This is given by $\psi = softmax(W_M^T\phi_2(x))$, where the output $\psi$ is a 3D tensor of size $T_m \times H_m \times W_m$. We further model $p(\mathcal{M}|x)$ by normalizing $\psi$ within each temporal slice: 
\begin{equation}
\small
\mathcal{M}_{m,n,t} = \frac{\psi_{m,n,t}}{\sum_{m, n}\psi_{m,n,t}},
\label{motorattention}
\end{equation}
where $\psi_{m,n,t}$ is the value at location $(m, n)$ and time step $t$ in the 3D tensor of $\psi$. And $\mathcal{M}$ can be considered as the expectation of $p(\mathcal{M}|x)$. 

\noindent \textbf{Stochastic Modeling}. Modeling motor attention in the context of forecasting human-object interaction requires a mechanism for addressing the stochastic nature of motor attention in developing the joint model. Here, we propose to use stochastic units to model the uncertainty. The key idea is to sample from the motor attention distribution. We follow the Gumbel-Softmax and reparameterization trick introduced in ~\cite{jang2016categorical,maddison2016concrete} to design a differentiable sampling mechanism:
\begin{equation}
\small
\label{eq:sample_motor}
\tilde{\mathcal{M}}_{m,n,t} \sim \frac{\exp ((\log \psi_{m,n,t} + G_{m,n,t})/\theta)}{\sum_{m,n} \exp ((\log \psi_{m,n,t} + G_{m,n,t})/\theta)},
\end{equation}
where $G$ is a Gumbel Distribution used to sample from discrete distribution. This Gumbel-Softmax trick produces a ``soft'' sampling step that allows the direct back-propagation of gradients to $\psi$. $\theta$ is the temperature parameter that controls the ``sharpness'' of the distribution. We set $\theta=2$ for all of our experiments.

\subsection{Interaction Hotspots Module}
The predicted motor attention $\mathcal{M}$ is further used to guide interaction hotspots estimation $p(\mathcal{A}|x)$ by considering the conditional probability
\begin{equation}
\label{eq:interaction}
\small
    p(\mathcal{A}|x) = \int_{\mathcal{M}} p(\mathcal{A}|\mathcal{M}, x) p(\mathcal{M}|x) d\mathcal{M}.
\end{equation}
In practice, $p(\mathcal{A}|x)$ is estimated using sampled motor attention $\tilde{\mathcal{M}}$ based on $p(\mathcal{A}|\tilde{\mathcal{M}}, x)$ and $p(\tilde{\mathcal{M}}|x)$. For each sample $\tilde{\mathcal{M}}$, $p(\mathcal{A}|\tilde{\mathcal{M}}, x)$ is defined by the interaction hotspots estimation function $F_A$. $F_A$ takes the input of a motor attention map $\tilde{\mathcal{M}}$ and $\phi_3(x)$, and has the form of a linear 2D convolution parameterzied by $W_A$ followed by a softmax function.
\begin{equation}
\label{eq:motorpooling}
\small
    p(\mathcal{A}|\tilde{\mathcal{M}}, x) = softmax\left(W_A^T (\tilde{\mathcal{M}} \otimes \phi_3(x) )\right),
\end{equation}
where $\otimes$ is the Hadamard product (element-wise multiplication). The result $p(\mathcal{A}|\mathcal{M}, x)$ is a 2D map of size $H_a \times W_a$. Intuitively, $\tilde{\mathcal{M}}$ presents a spatial-temporal saliency map to highlight feature representation $\phi_3(x)$. $F_A$ thus normalizes (using softmax) the output of a linear model on the selected features $\tilde{\mathcal{M}} \otimes \phi_3(x)$, and is a convex function. Finally, a similar sampling mechanism as in Eq.~\ref{eq:sample_motor} can be used to sample $\tilde{\mathcal{A}}$ from $p(\mathcal{A}|x)$.

\subsection{Anticipation Module}
We now present the last piece of our model---the action anticipation module. The action anticipation function $ p(y|\mathcal{A},\mathcal{M}, x) = F_P(\mathcal{A},\mathcal{M}, x)$ is defined as a function of the sampled motor attention map (3D) $\tilde{\mathcal{M}}$, sampled interaction heatmap (2D) $\tilde{\mathcal{A}}$ and the network feature $\phi_5(x)$. This is given by
\begin{equation}
\label{eq:predict}
\small
    p(y|\tilde{\mathcal{A}},\tilde{\mathcal{M}}, x) = softmax\left( W_P^T\Sigma 
    \left(\tilde{\mathcal{M}} \otimes \phi_5(x) \right)
     + W_P^T\Sigma\left(\tilde{\mathcal{A}} \odot \phi_5(x) \right)
    \right),
\end{equation}
where $\otimes$ is again the Hadamard product. $\Sigma$ is the global average pooling operation that pools a vector representation from a 2D or 3D feature map. $\odot$ is to use a 2D map ($\tilde{\mathcal{A}}$) to conduct Hadamard product to the last temporal slice of a 3D tensor $\phi_5(x)$. This is because the interaction hotspots $\tilde{\mathcal{A}}$ is only defined on the last observable frame. $W_P$ is a linear function that maps the features into prediction logits. $F_P$ is a combination of linear operations followed by a softmax function, and thus remains a convex function.

\subsection{Training and Inference}
Training our proposed joint model is challenging, as $p(\mathcal{M}|x)$ and $p(\mathcal{A}|\mathcal{M},x)$ are intractable. Fortunately, variational inference comes to the rescue.\\
\noindent\textbf{Prior Distribution}. During training, we assume that reference distributions of future hand position $Q({\mathcal{M}}|x)$ and interaction hotspots $Q({\mathcal{A}}|x)$ are known in prior. These distributions can be derived from manual annotation of 2D fingertips and interaction hotspots, as we will describe in Sec 4.1. A 2D isotropic Gaussian is further applied to the annotated 2D points, leading to the distributions of $Q({\mathcal{M}}|x)$ and $Q({\mathcal{A}}|x)$. If annotations are not available, we adopt uniform distributions for both $Q({\mathcal{M}}|x)$ and $Q({\mathcal{A}}|x)$.

\noindent\textbf{Variational Learning}. Our proposed model seeks to jointly predict motor attention $\mathcal{M}$, interaction hotspots $\mathcal{A}$, and the action label $y$. Therefore, we inject the posterior $p(\mathcal{A},\mathcal{M}|x)$ into $p(y|x)$. We further assume $p(\mathcal{A},\mathcal{M}|x)$ can be factorized into $p(\mathcal{A}|x)$ and $p(\mathcal{M}|x)$ (see supplementary materials for details). Our model thereby optimizes the resulting latent variable model by maximizing the Evidence Lower Bound (ELBO), given by\footnote{See supplementary material for the derivation.}
{\small
\begin{align}
%&\log p(y|x) \geq -\mathcal{L} \nonumber \\
\log p(y|x) \geq &E_{p(\mathcal{A},\mathcal{M}|x)}[\log p(y|\mathcal{A},\mathcal{M},x)] - log(p(\mathcal{A},\mathcal{M}|x))] \nonumber \\ 
%    = &\sum_{\mathcal{A},\mathcal{M}} \log p(y|\mathcal{A},\mathcal{M}, x) -KL[p(\mathcal{A},\mathcal{M}|x) ||Q(\mathcal{A},\mathcal{M}|x)] \nonumber \\ 
 = &\sum_{\mathcal{A},\mathcal{M}} \log p(y|\mathcal{A},\mathcal{M}, x) -KL[p(\mathcal{A}|x) ||Q(\mathcal{A}|x)] -KL[p(\mathcal{M}|x) ||Q(\mathcal{M}|x)].
\end{align}}%
Therefore, the loss function $\mathcal{L}$ is given by 
{\small
\begin{align}
\label{loss}
\mathcal{L} &=-\sum_{\mathcal{A},\mathcal{M}} \log p(y|\mathcal{A},\mathcal{M}, x) + KL[p(\mathcal{A}|x) ||Q(\mathcal{A}|x)] + KL[p(\mathcal{M}|x) ||Q(\mathcal{M}|x)].
\end{align}}%
The first term in the loss function is the cross entropy loss for action anticipation. The last two terms use KL-Divergence to align the predicted distributions of motor attention $p(\mathcal{M}|x)$ and interaction hotspots $p(\mathcal{A}|x)$ to their reference distributions ($Q(\mathcal{M}|x)$ and $Q(\mathcal{A}|x)$). To make the training practical, we draw a single sample for each input within a mini-batch similar to~\cite{jang2016categorical,maddison2016concrete}. Multiple samples of the same input will be drawn at different iterations. 

\noindent \textbf{Approximate Inference}. At inference time, our model could have drawn many samples of motor attention $\tilde{\mathcal{M}}$ and interaction hotspots $\tilde{\mathcal{A}}$ for the anticipation. However, the sampling and averaging is computationally expensive. We choose to feed deterministic $\mathcal{M}$ and $\mathcal{A}$ into Eq.\ \ref{eq:motorpooling} and Eq.\ \ref{eq:predict} at inference time. Note that $F_A$ and $F_P$ are convex, since they are composed of linear mapping function and softmax function. By Jensen's inequality, we have
\begin{equation}
\small
    E[F_A(\tilde{\mathcal{M}}, x)] \ge F_A(E[\tilde{\mathcal{M}}], x) = F_A(\mathcal{M}, x),
\end{equation}
\begin{equation}
\small
    E[F_P(\tilde{\mathcal{A}},\tilde{\mathcal{M}}, x)] \ge F_P(E[\tilde{\mathcal{A}}],E[\tilde{\mathcal{M}}], x) = F_P(\mathcal{A},\mathcal{M}, x)
\end{equation}
Therefore, such approximation provides a valid lower bound of $E[F_P(\tilde{\mathcal{A}},\tilde{\mathcal{M}}, x)]$ and $E[F_A(\tilde{\mathcal{M}}, x)]$, and serves as a shortcut to avoid sampling during testing.

\subsection{Network Architecture}
%\yin{We should separate out the network architecture and other implementation details (move to experiments). The architecture is more general and other details can be tied to an experiment. }
We consider two different backbone networks for our model, including lightweight I3D-Res50 network~\cite{carreira2017quo,wang2018non} pre-trained on Kinetics and heavy CSN-152~\cite{Tran_2019_ICCV} network pre-trained on IG-65M~\cite{ghadiyaram2019large}. We use I3D-Res50 for our ablation study on EGTEA and EPIC-Kitchens, and report results using CSN-152 backbone when competing on the EPIC-Kitchens dataset. Both networks have five convolutional blocks. The motor attention module, the interaction hotspots module and the recognition module are attached to the 2nd, the 3rd and the 5th block, respectively. We use 3D max pooling to match the size of attention map to the size of the feature map in Eq.\ \ref{eq:motorpooling} and Eq.\ \ref{eq:predict}. For training, our model takes an input of 32 frames (every other frame from a 64-frame chunk) with a resolution of $224\times224$. For inference, our model samples 30 clips from a video (3 along width of frame and 10 in time). Each clip has 32 frames with a resolution of $256\times256$. We average the scores of all sampled clips for video level prediction. Other implementation details will be discussed in the experiments.

\section{Experiments}
We now present our experiments and results. We briefly introduce our implementation details and describe the datasets and annotations. Moreover, we present our results on EPIC-Kitchens action anticipation challenge, followed by ablation studies that further evaluate our model on interaction hotspot estimation and motor attention prediction. Finally, we provide a discussion of our method.

\noindent \textbf{Implementation Details}. Our model is trained using SGD with momentum 0.9 and batch size 64 on 4 GPUs. The initial learning rate is 2.5e-4 with cosine decay. We set weight decay to 1e-4 and enable batch norm~\cite{ioffe2009batch}. We downsample all frames to 320x256 (24fps) for EGTEA, and 512x288 (30fps) for EPIC-Kitchens. We apply several data augmentation techniques, including random flipping, rotation, cropping and color jittering to avoid overfitting.

%We start with an introduction the dataset, evaluation metrics, and implementation details. We then present our experiments results. Firstly, we demonstrate our results on EPIC-Kitchens action anticipation challenge. To analysis the role of each components in our model, we then provide a comprehensive ablation study of our model. We also present our results on interaction hotspots estimation and motor attention prediction, and compare to a set of strong baselines. Finally, we provide insightful analysis and discussion of our method.

\subsection{Datasets and Annotations}
\noindent \textbf{Datasets}. We make use of two FPV datasets: EGTEA Gaze+~\cite{Li_2018_ECCV,li2020eye} and Epic-Kitchens~\cite{Damen2018EPICKITCHENS}. EGTEA comes with $10,321$ action instances from $19/53/106$ verb/noun/action classes. We report results on the first split of the dataset. EPIC-Kitchens contains $39,596$ instances from 125 verbs and 352 nouns. We follow~\cite{furnari2019rulstm} to split the public training set into training and validation sets with 2513 action classes. We conduct ablation studies on this train/val split, and present the action anticipation results on the testing sets.
We set the anticipation time as $0.5$ seconds for EGTEA and $1$ second~\cite{Damen2018EPICKITCHENS} for EPIC-Kitchens.

%For the EGTEA, we set anticipation time as $0.5$ seconds. For the EPIC-Kitchens, we set anticipation time as $1$ second as defined in the Anticipation Challenge. 

\noindent \textbf{Annotations}. Our model requires supervisory signals of interaction hotspots and hand trajectories during training. We provide extra annotations for both EGTEA and EPIC-Kitchens datasets. These annotations will be made publicly available. Specifically, we manually annotated interaction hotspots as 2D points on the last observable frames for all instances on EGTEA and a subset of instances on EPIC-Kitchens. This is because many noun labels in Epic-Kitchens have very few instances, hence we focus on interaction hotspots of action instances that include many-shot nouns~\cite{Damen2018EPICKITCHENS} in the training set. 

Moreover, we explore different approaches to generate the pseudo ground truth of future hand trajectories. On EGTEA, we trained a hand segmentation model (\cite{long2015fully} using hand masks from the dataset). The motor attention was approximated by segmenting hands at every frame and tracking the fingertip closest to an active object. To mitigate ego-motion, we used optical flow and RANSAC to compute a homography transform, and project the motor attention to the last observable frame. As EPIC-Kitchens does not provide hand masks, we instead annotated the fingertip closest to an interaction hotspots on the last observable frame. A linear interpolation of 2D motion between the fingertip and the interaction hotspots was used to approximate the motor attention.

% \noindent \textbf{Implementation Details}. 
% We downsampled all frames to $320\times256$ with 24 fps for the EGTEA dataset, and $512\times288$ with 30 fps for the EPIC-Kitchens dataset. For training, we applied several data augmentation techniques, including random flipping, rotation, cropping and color jittering to avoid overfitting. Our model was trained with cross entropy loss using SGD with momentum of 0.9. The batch size was 64/16 distributed on 4 GPUs (I3D-Res50/CSN-152). Synchronized batch normalization was enabled. The initial learning rate was 2.5e-4 (linear rescaled for smaller batch size) with cosine decay. I3D models were trained for $40$/$30$ epochs on EGTEA/EPIC-Kitchens, while CSN-152 models were only trained for 18 epochs for competing on EPIC-Kitchens challenge. 
\subsection{FPV Action Anticipation on EPIC-Kitchens}
\begin{table}[t]
\footnotesize 
% \footnotesize

\centering
\caption{Action anticipation results on Epic-Kitchens. Ours+Obj model outperforms state-of-the-art by a notable margin. See discussions of Ours+Obj in Sec.\ 4.2.}
{
\tablestyle{2pt}{1.0}
\setlength{\tabcolsep}{5pt} 
\renewcommand{\arraystretch}{1.0} % Default value: 1
\begin{tabular}{c|c|ccc}
\multicolumn{1}{c}{\multirow{2}{*}{}}  
&\multicolumn{1}{c|}{\multirow{2}{*}{Method}}                                        
& \multicolumn{3}{c}{Top1/Top5 Accuracy} \\
\multicolumn{2}{c|}{}   & Verb & Noun &Action         \\ \hline 
\multirow{7}{*}{s1}  
&\makecell{2SCNN~\cite{Damen2018EPICKITCHENS}}   & 29.76 / 76.03 & 15.15 / 38.65 &4.32 / 15.21 \\
&\makecell{TSN~\cite{Damen2018EPICKITCHENS}}      & 31.81 / 76.56 & 16.22 / 42.15 &6.00 / 18.21  \\
% &\makecell{TSN Flow~\cite{wang2016temporal}}     & 29.64/73.70 & 10.30/30.09 &2.93/10.92   \\
% &\makecell{TSN Fusion~\cite{wang2016temporal}}         & 30.66/75.32 & 14.86/40.11 &4.62/16.01  \\	
&\makecell{TSN+MCE~\cite{Furnari_2018_ECCV_Workshops}}     & 27.92 / 73.59 & 16.09 / 39.32 & 10.76 / 25.28  \\ 
&\makecell{Trans R(2+1)D~\cite{Miech_2019_CVPR_Workshops}}   & 30.74 / 76.21 & 16.47 / 42.72 &9.74 / 25.44   \\ 
&\makecell{RULSTM~\cite{furnari2019rulstm}}     & 33.04 / \textbf{79.55} & 22.78 / 50.95 &14.39 / 33.73   \\ 
&\makecell{Ours}      & 34.99 / 77.05 & 20.86 / 46.45 &14.04 / 31.29   \\ 
&\makecell{Ours+Obj}      & \textbf{36.25} / 79.15 & \textbf{23.83} / \textbf{51.98} &\textbf{15.42} / \textbf{34.29}   \\ \hline
\multirow{7}{*}{s2}  
&\makecell{2SCNN~\cite{Damen2018EPICKITCHENS}}   & 25.23 / 68.66 & 9.97 / 27.38 &2.29 / 9.35 \\
&\makecell{TSN~\cite{Damen2018EPICKITCHENS}}      & 25.30 / 68.32 & 10.41 / 29.50 &2.39 / 9.63  \\
% &\makecell{TSN Flow~\cite{wang2016temporal}}     & 25.61/67.57 & 8.40/24.62 &1.78/8.19   \\
% &\makecell{TSN Fusion~\cite{wang2016temporal}}         & 25.37/68.25 & 9.76/27.24 & 1.74/9.05  \\	
&\makecell{TSN+MCE~\cite{Furnari_2018_ECCV_Workshops}}     & 21.27 / 63.66 & 9.90 / 25.50 & 5.57 / 25.28  \\ 
&\makecell{Trans R(2+1)D~\cite{Miech_2019_CVPR_Workshops}}     &28.37 / 69.96 & 12.43 / 32.20 &7.24 / 19.29  \\ 
&\makecell{RULSTM~\cite{furnari2019rulstm}}     & 27.01 / 69.55 & 15.19 / 34.38 & 8.16 / 21.20  \\ 
&\makecell{Ours}     & 28.27 / 70.67 & 14.07 / 34.35 &8.64 / 22.91  \\ 
&\makecell{Ours+Obj}      & \textbf{29.87} / \textbf{71.77} & \textbf{16.80} / \textbf{38.96} &\textbf{9.94} / \textbf{23.69}   \\ 
\end{tabular}}

\label{table:actionanticipation}
\end{table}

We highlight our results for FPV action anticipation on EPIC-Kitchens dataset. 

\noindent \textbf{Experiment Setup}. To compete for EPIC-Kitchens anticipation challenge, we used the backbone network CSN152. We trained our model on the public training set and report results using top-1/5 accuracy as in~\cite{Damen2018EPICKITCHENS}.

\noindent \textbf{Results}. Table~\ref{table:actionanticipation} compares our results to latest methods on EPIC-Kitchens. Our model outperforms strong baselines (TSN and 2SCNN) reported in~\cite{Damen2018EPICKITCHENS} by a large margin. Compared to previous best results from RULSTM~\cite{furnari2019rulstm}, our model archives +2\%/-1.9\%/-0.3\% for verb/noun/action on seen set, and +1.3\%/-1.1\%/+0.6\% on unseen set of EPIC-Kitchens. Our results are better for verb, worse for noun and comparable or better for actions. Notably, RULSTM requires object boxes \& optical flow for training and object features \& optical flow for testing. In contrast, our method uses hand trajectories and interaction hotspots for training and needs \emph{only RGB frames} for testing. 

To further improve the performance, we fuse the object stream from RULSTM with our model (Ours+Obj). Compared to RULSTM, Ours+Obj has a performance gain of $+3.2\%$/$+2.9\%$ for verb, $+1.1\%$/$+1.6\%$ for noun, and $+1.0\%$/$+1.8\%$ for action (seen/unseen). It is worthy pointing out that RULSTM benefits from an extra flow network, while ours+Obj model takes additional supervisory signals of hands and hotspots. Note that our performance boost does not simply come from those extra annotations. In a subsequent ablation study, we have shown that simply training with these extra annotations has minor improvement, when used without our proposed probabilistic deep model. 

We note that it is not possible to make a direct apples-to-apples comparison between our model and RULSTM~\cite{furnari2019rulstm}, as the two models used vastly different training signals. We refer readers to the supplementary materials for a detailed experiment setup comparison. In terms of performance, our model is comparable to RULSTM without using any side information for inference. When using additional object stream during inference as in RULSTM, our model outperforms RULSTM by a relative improvement of \textbf{7\%/22\%} on seen/unseen set. More importantly, our model also provides the additional capabilities of predicting future hand trajectories and estimating interaction hotspots.

\subsection{Ablation Study}
We present ablation studies of our model. We introduce our experiment setup, evaluate each component of our model, and then contrast our method to a series of baselines on motor attention prediction and interaction hotspot estimation

\begin{table}[t]
\footnotesize 
\caption{Ablation study for action anticipation. We compare our model with backbone I3D network, and further analyze the role of motor attention prediction, interaction hotspots estimation, and stochastic units in joint modelling. See discussions in Sec.\ 4.3.}
\centering
{
\setlength{\tabcolsep}{2.7pt} % Default value: 6pt
\renewcommand{\arraystretch}{1.05} % Default value: 1
\tablestyle{2pt}{1.0}
\scriptsize
\begin{tabular}{c|ccc|ccc}
\multicolumn{1}{c|}{\multirow{3}{*}{Method}}          
&\multicolumn{3}{c|}{EGTEA} &\multicolumn{3}{c}{Epic-Kitchens} \\ \cline{2-7}
&\multicolumn{3}{c|}{Top1 Accuracy / Mean Cls Accuracy} &\multicolumn{3}{c}{Top1 Accuracy / Top5 Accuracy} \\
\multicolumn{1}{c|}{}   & Verb & Noun  & Action      & Verb & Noun  & Action    \\ \hline 

\makecell{I3D-Res50}   & 48.01/31.25& 42.11/30.01 & 34.82/23.20 & 30.06/76.86 & 16.07/41.67  & 9.60/24.29\\
%\makecell{SoftAtten}      & 48.09/31.35   & 42.3/30.28  & 35.03/23.51 & 29.75/75.45 & 15.95/42.01  & 9.53/24.07\\

\makecell{JointDet}         & 48.58/32.21   & 43.95/31.26  & 35.69/23.59  & 30.16/\textbf{76.86}   & 16.25/41.71  &9.76/24.40\\	
\makecell{Hotspots Only}      &47.95/31.94    & 44.02/32.53  &35.50/23.82 &30.21/75.93  &16.57/42.28 &9.66/24.33   \\ 
\makecell{Motor Only}      &\textbf{49.35}/32.34 &\textbf{45.69}/\textbf{33.93} &36.49/25.13 
& 30.63/76.69   & 17.28/42.56  &10.21/25.32\\ 
\makecell{Ours}      &48.96/\textbf{32.48}    &45.50/32.73 &\textbf{36.60}/\textbf{25.30}
& \textbf{30.65}/76.53 &\textbf{17.40}/\textbf{42.60}  &\textbf{10.38}/\textbf{25.48}\\ 
\end{tabular}}

\label{table:ablation:action}
\end{table}

\begin{table}[t]
\caption{Ablation study for interaction hotspots estimation. Jointly modeling motor attention with stochastic units can greatly benefit the performance of interaction hotspots estimation. ($\uparrow$/$\downarrow$ indicates higher/lower is better) See discussions in Sec.\ 4.3.}
\centering
{
\tablestyle{2pt}{1.0}
\setlength{\tabcolsep}{3pt} 
\renewcommand{\arraystretch}{1.05}
\begin{tabular}{c|cccc|cccc}
\multicolumn{1}{c|}{\multirow{3}{*}{Method}}          
&\multicolumn{4}{c|}{EGTEA} &\multicolumn{4}{c}{Epic-Kitchens} \\ \cline{2-9}
\multicolumn{1}{c|}{}   & Prec $\uparrow$ & Recall $\uparrow$  & F1 $\uparrow$  &KLD $\downarrow$       & Prec $\uparrow$ & Recall $\uparrow$  & F1 $\uparrow$  &KLD $\downarrow$  \\ \hline 

\makecell{I3DHeatmap}   & 12.82 & 37.53 & 19.11 &2.66 &17.20 &77.39 & 28.15 &3.07\\
\makecell{JointDet}         & 16.11  & 41.82  & 23.26 &1.84  &17.32 &85.79 &28.83 &2.21\\													               
\makecell{Ours}     &\textbf{17.43}     &\textbf{48.81} &\textbf{25.69} &\textbf{1.62} &\textbf{17.86} &\textbf{86.59} &\textbf{29.60} &\textbf{1.99}\\ 
\end{tabular}}

\label{table:ablation:hotspots}
\end{table}

\noindent \textbf{Experiment Setup}. For all of our ablation studies, we adopt the lightweight I3D-Res50~\cite{wang2018non} as backbone network to reduce computational cost. Our model is evaluated for action anticipation, motor attention prediction and interaction hotspots estimation across EGTEA (using split1) and EPIC-Kitchens (using the train/val split from~\cite{furnari2019rulstm}). Specifically, we consider the following metrics.
\begin{itemize}[leftmargin=*]
    \item \textbf{Action Anticipation}. We report Top1/Mean Class accuracy on EGTEA as in~\cite{li2015delving} and Top1/Top5 accuracy as on EPIC-Kitchens following~\cite{furnari2019rulstm}.
    \item \textbf{Interaction Hotspots Estimation}. We report F1 score as in~\cite{Li_2018_ECCV} and KL-Divergence (KLD) as in~\cite{nagarajan2018grounded} using a downsampled heatmap (32x) at the last observable frame. 
    \item \textbf{Motor Attention Prediction}. We report the average and final displacement errors between the most confident location on a predicted attention map and the ground-truth hand points, similar to previous work on trajectory prediction~\cite{alahi2016social}. Note that the motor attention maps is downsampled by a factor of 32/8 in space/time. Hence, we report displacement errors normalized in spatial and temporal dimension.
\end{itemize}

\noindent \textbf{Benefits of Joint Modeling}. As a starting point, we compare our model with a backbone I3D-Res50 model. We present the results of action anticipation in Table~\ref{table:ablation:action}. In comparison to I3D-Res50, our model improves noun and action prediction by $+3.4\% / 1.8\%$ on EGTEA and $+1.3\% / 0.8\%$ on EPIC-Kitchens. Moreover, we show that our model improves the performance of interaction hotspots estimation. We consider the baseline I3D model that only estimates interaction region with interaction hotspots module as I3DHeatmap. As shown in Table~\ref{table:ablation:hotspots}, our model improves the F1 score by $6.6\% / 1.5\% $ on EGTEA/EPIC-Kitchens. 

\noindent \textbf{Stochastic Modeling vs.\ Deterministic Modeling}. We further evaluate the benefits of probabilistic modeling of motor attention and interaction hotspots. To this end, we compare our model with a deterministic joint model (\emph{JointDet}). JointDet has the same architecture as our model, except for the stochastic units. As shown in Table~\ref{table:ablation:action}, JointDet slightly improve the I3D baseline for action anticipation ($+0.87\%$ on EGTEA and $+0.16\%$ on EPIC-Kitchens), yet lags behind our probabilistic model. Specifically, our model outperforms JointDet by $0.91\%$ and $0.62\%$ on EGTEA and EPIC-Kitchens. Moreover, in comparison to JointDet, our model has better performance for interaction hotspots estimation ($+2.4\% / +0.8\%$ in F1 scores on EGTEA/EPIC-Kitchens). These results suggest that simply training with extra annotations might fail to capture the uncertainty of visual anticipation. In contrast, our design choice of probabilistic modeling can effectively deal with those uncertainty, therefore helps to improve the performance of joint modeling.

\noindent \textbf{Motor Attention vs.\ Interaction Hotspots}. Futhermore, we evaluate the contributions of motor attention and interaction hotspots for FPV action anticipation. We consider two baseline models in Table~\ref{table:ablation:hotspots}: I3D model equipped with only motor attention module (\emph{Motor Only}), and I3D model equipped with only interaction hotspots module (\emph{Hotspots Only}). Both models underperform the full model across the two datasets, yet the gap between \emph{Motor Only} and the full model is smaller. These results suggest that both components contribute to the performance boost of action anticipation, yet the modeling of motor attention weights more than the modeling of interaction hotspots.

\noindent \textbf{Interaction Hotspots Estimation}.
We present additional results on interaction hotspots estimation. We compare our results to the following baselines. 
\begin{itemize}[leftmargin=*]
    \item {\bf Center Prior} represents a Gaussian Distribution at the center of the image.
    \item {\bf Grad-Cam} uses the same I3D backbone network as our model, and produces a saliency map via Grad-Cam~\cite{selvaraju2017grad}.
    \item {\bf EgoGaze} considers possible gaze position as salient region of a given image. This model is trained on eye fixation annotation from EGTEA-Gaze+~\cite{huang2018predicting}. The assumption is that the person is likely to look at the interaction hotspots.
    \item {\bf DSS Saliency} predicts salient region during human object interaction. This model is trained on pixel-level saliency annotation from~\cite{hou2017deeply}.
    \item {\bf EgoHotspots} is the latest work~\cite{nagarajan2018grounded} for estimating interaction hotspots. 
\end{itemize}

\begin{table}[t]
\caption{Interaction hotspots estimation results on EGTEA and EPIC-Kitchens. Our model outperforms a set of strong baselines. ($\uparrow$/$\downarrow$ indicates higher/lower is better)}
\tablestyle{2pt}{1.0}
\setlength{\tabcolsep}{5.0pt} 
\scriptsize
 % Default value: 1
\begin{tabular}{c|cccc|cccc}
\multicolumn{1}{c|}{\multirow{3}{*}{Method}}          
&\multicolumn{4}{c|}{EGTEA} &\multicolumn{4}{c}{Epic-Kitchens} \\ \cline{2-9}
\multicolumn{1}{c|}{}   & Prec$\uparrow$ & Recall$\uparrow$  & F1$\uparrow$  &KLD$\downarrow$       & Prec$\uparrow$ &Recall$\uparrow$  & F1$\uparrow$  &KLD$\downarrow$  \\ \hline 
\makecell{Center Prior}     &10.87    &17.65 &13.45 &10.64 &11.66 & 16.97 & 13.82 &10.27 \\
\makecell{Grad-Cam~\cite{selvaraju2017grad}}     & 9.98 &22.13   &13.76 &8.73 &10.85 &20.01 & 14.07 & 8.06 \\ 
\makecell{DSS~\cite{hou2017deeply}}     & 9.02 &39.49   &14.69 &6.12 &12.03 &33.75 & 17.74 & 5.21 \\ 
\makecell{EgoGaze~\cite{huang2018predicting}}     & 15.02 &31.34   &20.31 &3.20 &11.30 &27.65 & 16.05 & 3.37 \\ 
\makecell{EgoHotspots~\cite{nagarajan2018grounded}}     & 16.51 &24.07   &19.59 &3.36 &\textbf{22.26} &31.37 &26.04 &2.84 \\ 
\makecell{Ours}     &\textbf{17.43}     &\textbf{48.81} &\textbf{25.69} &\textbf{1.62}  &17.86 & \textbf{86.5} & \textbf{29.6} &\textbf{1.99}\\ 
\end{tabular}

\label{table:hotspots}
\end{table}

Our results are shown in Table~\ref{table:hotspots}. Our model outperforms the best baselines (EgoGaze and EgoHotspots) by $5.4\%$ on EGTEA and $3.6\%$ on EPIC-Kitchens in F1 scores. These results suggest that our proposed joint model can effectively identify future interaction region. Another observation is that our model performs better on EPIC-Kitchens than EGTEA. This is probably due to the larger number of available training samples.

% These are weak arguments
%This is because the many-shot nouns in EPIC-Kitchens are composed mostly of rigid objects. In contrast, non-rigid objects (tomato, lettuce, cucumber etc.) take up a large proportion of the EGTEA dataset, This incurs additional challenge to estimating interaction hotspots, since the interaction region changes dramatically due to deformation.\\

\begin{table}[t]
\caption{Motor attention prediction results on EGTEA. Our model compares favourably to strong baselines. ($\uparrow$/$\downarrow$ indicates higher/lower is better)}
\label{table:motor_attention}
\centering
\begin{tabular}{c|c|c}
Method                                  & Avg. Disp. Error  $\downarrow$        & Final Disp. Error  $\downarrow$      \\ \hline 
Kalman Filter                              & 0.32       & 0.48    \\ 
GPR                        & 0.29       & 0.37    \\ 
LSTM                              & \textbf{0.22}       & \textbf{0.35}    \\ 
Ours                             & 0.23       & 0.36    \\ 
\end{tabular}
\end{table}

\noindent \textbf{Motor Attention Prediction}.
We report our results on motor attention prediction. We consider the following baselines and only report results on EGTEA, as the future hand position on EPIC-Kitchens is not accurate (see Sec.\ 4.1).
\begin{itemize}[leftmargin=*]
    \item {\bf Kalman Filter} describes the hand trajectory prediction problem with state-space model, and assumes linear acceleration during update step.
    \item {\bf Gaussian Process Regression (GPR)} iteratively predicts the future hand position using Gaussian Process Regression. 
    \item  {\bf LSTM} adopts a vanilla LSTM network for trajectory forecasting. We use the implementation from~\cite{alahi2016social}.
\end{itemize}

\begin{figure*}[t]
\centering
\includegraphics[width=0.9\linewidth]{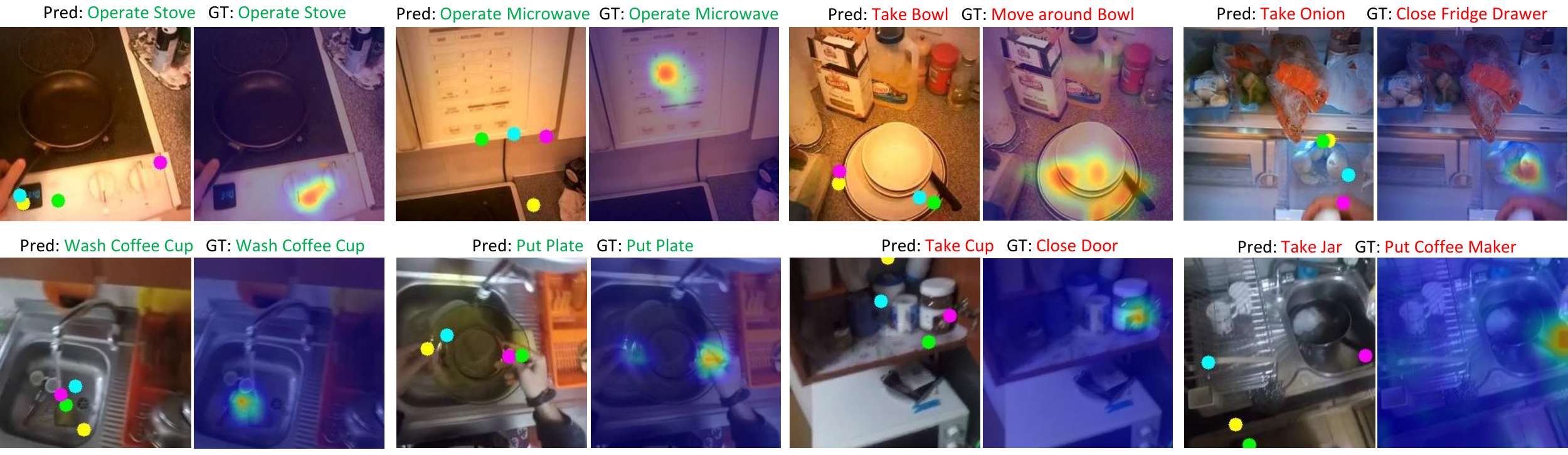}
\caption{Visualization of motor attention (left image), interaction hotspots (right image), and action labels (captions above the images) on sample frames from EGTEA (first row) and EPIC-Kitchens (second row). Both successful ({\color{green} green} label) and failure cases ({\color{red} red} label) are shown. Future hands position are predicted at every 8 frames and plotted on the last observable frame with the order of {\color{yellow} yellow}, {\color{green} green}, {\color{cyan} cyan}, and {\color{magenta} magenta}.} 
\label{fig:vis}
\end{figure*}

The results are presented in Table~\ref{table:motor_attention}. Our model outperforms Kalman filter and GPR, yet is slightly worse than LSTM model (+$0.01$ in both errors). Note that all baseline methods need the coordinate of the first observed hand for prediction. This simplifies trajectory prediction into a less challenging regression problem. In contrast, our model does not need hand coordinates for inference. A model that relies on the observation of hand positions will encounter failure cases when the hand has not been observed, while our model is still capable of ``imagining'' the possible hand trajectory. See ``Operate Microwave'' and ``Wash Coffee Cup'' in Fig.\ \ref{fig:vis} for example results from our model. %This generalization ability is attributable to our incorporation of a probabilistic model of motor attention. Note that motor attention here serves a vehicle towards learning the future representation.\\

\noindent \textbf{Visualization of Motor Attention and Interaction Hotspots}. Finally, we visualize the predicted motor attention, interaction hotspots, and action labels from our model in Fig.\ \ref{fig:vis}. The predicted motor attention almost always attends to the predicted objects and corresponding interaction hotspots. Hence, our model can address challenging cases where next-active objects are ambiguous. Take the first example of ``Operate Stove'' in Fig.\ \ref{fig:vis}. Our model successfully predicted the future objects and estimated the interaction hotspots as the stove control knob.\\

\subsection{Remarks and Discussion}
%To summarize, our method achieves new state-of-the-art results for FPV action anticipation. Through extensive ablation studies, we demonstrate (1) the benefit of joint modeling; (2) the advantage of probabilistic modeling; (2) the contribution of motor attention and interaction hotspots for action anticipation; and (3) the effectiveness of our model in the tasks of motor attention prediction and interaction hotspots estimation. 

We must also point out that our method has certain limitations, which point to exciting future research directions. For example, our model requires additional annotations for training, which might bring scalability issues when analyzing other datasets. These dense annotations can indeed be approximated using sparsely annotated frames as discussed in Sec.\ 4.1. We speculate that more advanced hand tracking and object segmentation models can be explored to generating the pseudo ground truth of motor attention and interaction hotspots. Moreover, our model shares a similar conundrum faced by previous work on anticipation. Our model is likely to fail when future active objects are not observed. See ``Close Fridge Drawer''and ``Put Coffee Maker'' in Fig.\ \ref{fig:vis}. We conjecture that these cases requires incorporating logical reasoning into learning based methods---an active research topic in our community.

% This connects to a more general question in Artificial Intelligence: How can we endow an intelligent system with the ability of exploration and logical reasoning? The solution remains to be explored. 

% We have shown extensive quantitative and qualitative results to demonstrate the strength of our method. However, our method also subjects to certain limitations, which point to exciting future research directions. Our model requires additional video annotations for training, which might bring scalability issues when applying to other datasets. These dense annotations can indeed be approximated using sparsely annotated frames as discussed in Sec.\ 4.1. We also speculate that these annotations can be extracted using more advanced hand tracking and object part segmentation models. Moreover, our model shares a similar conundrum faced by previous anticipation studies. The model will fail when future active objects are occluded or not even observed. (See ``Close Fridge Drawer''and ``Put Coffee Maker'' in Fig.\ \ref{fig:vis}) This connects to a more general question in Artificial Intelligence: How can we endow an intelligent system with the ability of exploration and logical reasoning? The solution remains to be explored. 
\section{Conclusions}
We presented the first deep model that jointly predicts motor attention, interaction hotspots, and future action labels in FPV. Importantly, we demonstrated that motor attention plays an important role in forecasting human-object interactions. Another key insight is that characterizing motor attention and interaction hotspots as probabilistic variables can account for the stochastic pattern of human intentional movement. We believe that our model provides a solid step towards the challenging problem of visual anticipation. 
%. Our model achieved new state-of-the-art action anticipation results on two FPV benchmarks datasets, and obtained strong results on motor attention prediction and interaction hotspots estimation.

\noindent \textbf{Acknowledgments}. Portions of this research were supported in part by National Science Foundation Award 1936970 and a gift from Facebook. YL acknowledges the support from the Wisconsin Alumni Research Foundation.

\bibliographystyle{splncs04}
\bibliography{egbib}

This is the supplementary material for our submission to ECCV 2020, titled ``Forecasting Human-Object Interaction: Joint Prediction of Motor Attention and Actions in First Person Video''. 
%We hope this document will complement to our submission. 
The contents are organized as follows.
\begin{itemize}
\item \hyperref[sec:s1]{A} Network Architecture.
\item \hyperref[sec:s2]{B} Mathematical Derivation for Equation 8.
\item \hyperref[sec:s3]{C} Details on Data Annotation.
\item \hyperref[sec:s4]{D} Experiment Setup Comparison to RULSTM.
\item \hyperref[sec:s5]{E} Epic-Kitchens Challenge Leaderboard.
\item \hyperref[sec:s6]{F} Experiments on Gaze Fixation Model.
\item \hyperref[sec:s7]{G} Additional Qualitative Results.
\end{itemize}

\renewcommand\thesection{\Alph {section}}
\renewcommand\thesubsection{\thesection.\arabic{subsection}}

\section{Network Architecture}
\label{sec:s1}

\noindent \textbf{Network Architecture}. We present our network architecture using 3D Res50 backbone in Table~\ref{table:structure}. A similar architecture is also used for CSN-152 backbone. We followed ~\cite{Li_2018_ECCV} to use the features from bottom layers of the network for motor attention prediction and interaction hotspots estimation, and the features from the top layer for action anticipation. Our model jointly predicts motor attention, interaction hotspots and future actions, and thus is conceptually similar to multi-task learning e.g.,~\cite{misra2016cross}. The key difference is that outputs of our model depends on each other. For example, motor attention is used for interaction hotspots estimation and both motor attention and interaction hotspots are used for action anticipation.

\section{Mathematical Derivation for Equation 7}
\label{sec:s2}
We present the derivation of our variational learning as discussed in Sec 3.5. Specifically, we inject posterior $p(\mathcal{A},\mathcal{M}|x)$ into $p(y|x)$ and optimize the resulting latent variable model by maximizing the Evidence Lower Bound (ELBO). However, the prior distribution of $Q(\mathcal{A},\mathcal{M}|x)$ is not available for training. Hence, we further approximate $p(\mathcal{A},\mathcal{M}|x)$ by factorizing it into $p(\mathcal{A}|x)$ and $p(\mathcal{M}|x)$. Namely, we assume that $\mathcal{A}$ and $\mathcal{M}$ is conditionally independent given the input $x$. Thus, we have
%$\mathcal{A}$ and $\mathcal{M}$ are independent when $\mathcal{M}$ is given
{\small
\begin{align}
&KL[p(\mathcal{A},\mathcal{M}|x) ||Q(\mathcal{A},\mathcal{M}|x)] \nonumber \\ 
= & KL[p(\mathcal{A}|\mathcal{M},x) ||Q(\mathcal{A}|\mathcal{M},x)]+ KL[p(\mathcal{M}|x) ||Q(\mathcal{M}|x)]. \nonumber
\end{align}}%
The ELBO of our proposed joint model can be derived as
{\small
\begin{align}
\log p(y|x) \geq \ & E_{p(\mathcal{A},\mathcal{M}|x)}[\log p(y|\mathcal{A},\mathcal{M},x)] - log(p(\mathcal{A},\mathcal{M}|x))] \nonumber \\ 
    = \ &\sum_{\mathcal{A},\mathcal{M}} \log p(y|\mathcal{A},\mathcal{M}, x) -KL[p(\mathcal{A},\mathcal{M}|x) ||Q(\mathcal{A},\mathcal{M}|x)] \nonumber \\ 
 = \ &\sum_{\mathcal{A},\mathcal{M}} \log p(y|\mathcal{A},\mathcal{M}, x) -KL[p(\mathcal{A}|x) ||Q(\mathcal{A}|x)] -KL[p(\mathcal{M}|x) ||Q(\mathcal{M}|x)]. \nonumber
\end{align}}%

\section{Details on Data Annotation}
\begin{figure}[t]
\centering
\includegraphics[width=0.98\linewidth]{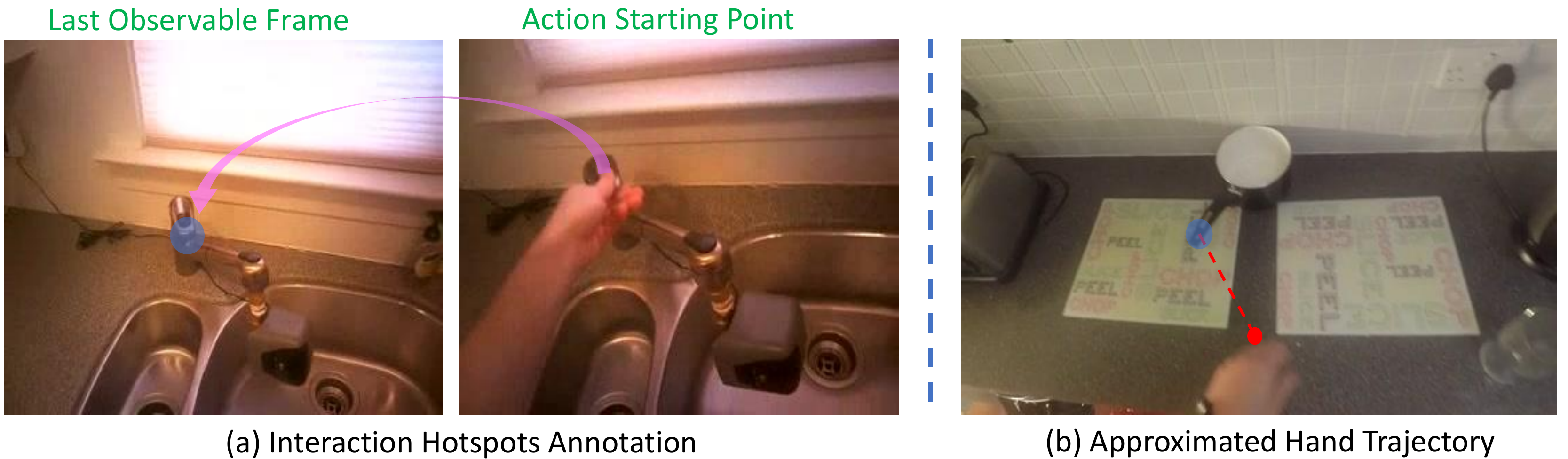}
\caption{(a) illustrates the interaction hotspots annotation process. (b) illustrates the approximation of the future hand trajectory on the Epic-Kitchens dataset.} 
\label{fig:data}
\end{figure}

\label{sec:s3} 
We provide additional details on data annotation. In Sec.\ 4.1, we introduced how we obtain the prior distribution of motor attention and interaction htospots. Here we provide a visual illustration of our efforts on the data annotation. As shown in Fig.~\ref{fig:data} (a), we compare the last observable frame with the first frame of action segment. If the active object presents in the last observable frame, we annotate the corresponding contact point and enforce a 2D Gaussian distribution to imitate the stochastic patterns of human-object interaction. Since the hand mask is absent from EPIC-Kitchens dataset, we adopt a 2D interpolation between the the finger tip annotation and interaction hotspots annotation to generate the pseudo ground truth of future hand trajectory (Take Fig.~\ref{fig:data} (b) for an instance). Note that we use a smaller anticipation time (0.5s) on the EGTEA dataset. This is because the EGTEA dataset has a smaller angle of view in comparison with the EPIC-Kitchens dataset. A large anticipation time will reduce the number of samples that have next-active objects on the last observable frame. To summarize, there are $14951$ annotated sample on the EPIC-Kitchens Dataset, and $7381$ annotated samples on the EGTEA dataset. We believe those additional annotations can facilitate future research of human-object interaction in FPV.

\begin{figure*}[t]
\centering
\includegraphics[width=1\linewidth]{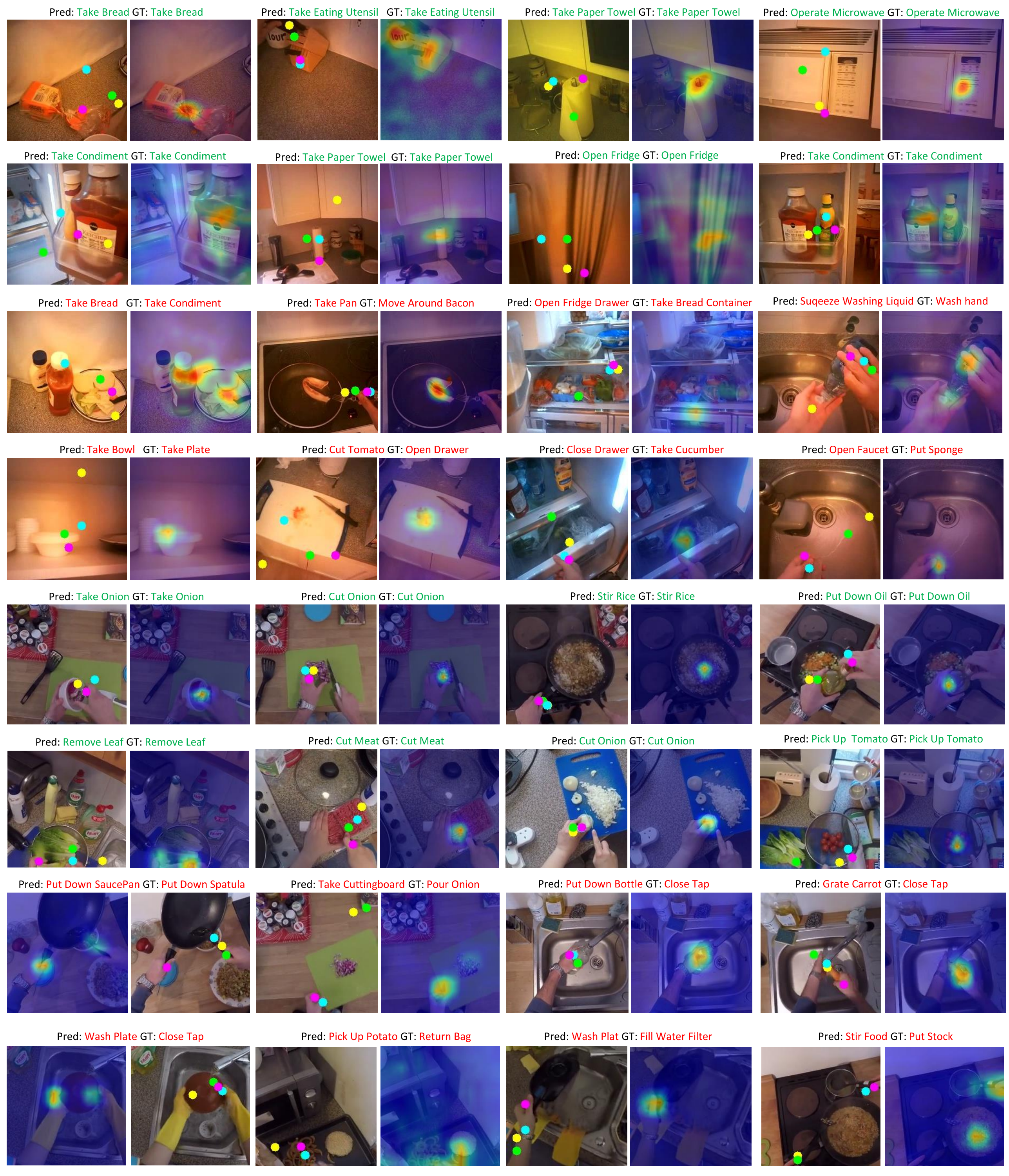}
\caption{Additional visualization of predicted motor attention (left), interaction hotspots (right), and action labels (top) from the EGTEA (row 1-4) and EPIC-Kitchens (row 5-8). Both successful cases ({\color{green} green} label) and failure cases ({\color{red} red} label) are presented. Future hands position are downsampled by a temporal factor of 8, and forecasted to the last observable frame in the order of {\color{yellow} yellow} , {\color{green} green}, {\color{cyan} cyan}, and {\color{magenta} magenta}.} 
\label{fig:vis-more}
\end{figure*}

\begin{figure*}[t]
\centering
\includegraphics[width=1\linewidth]{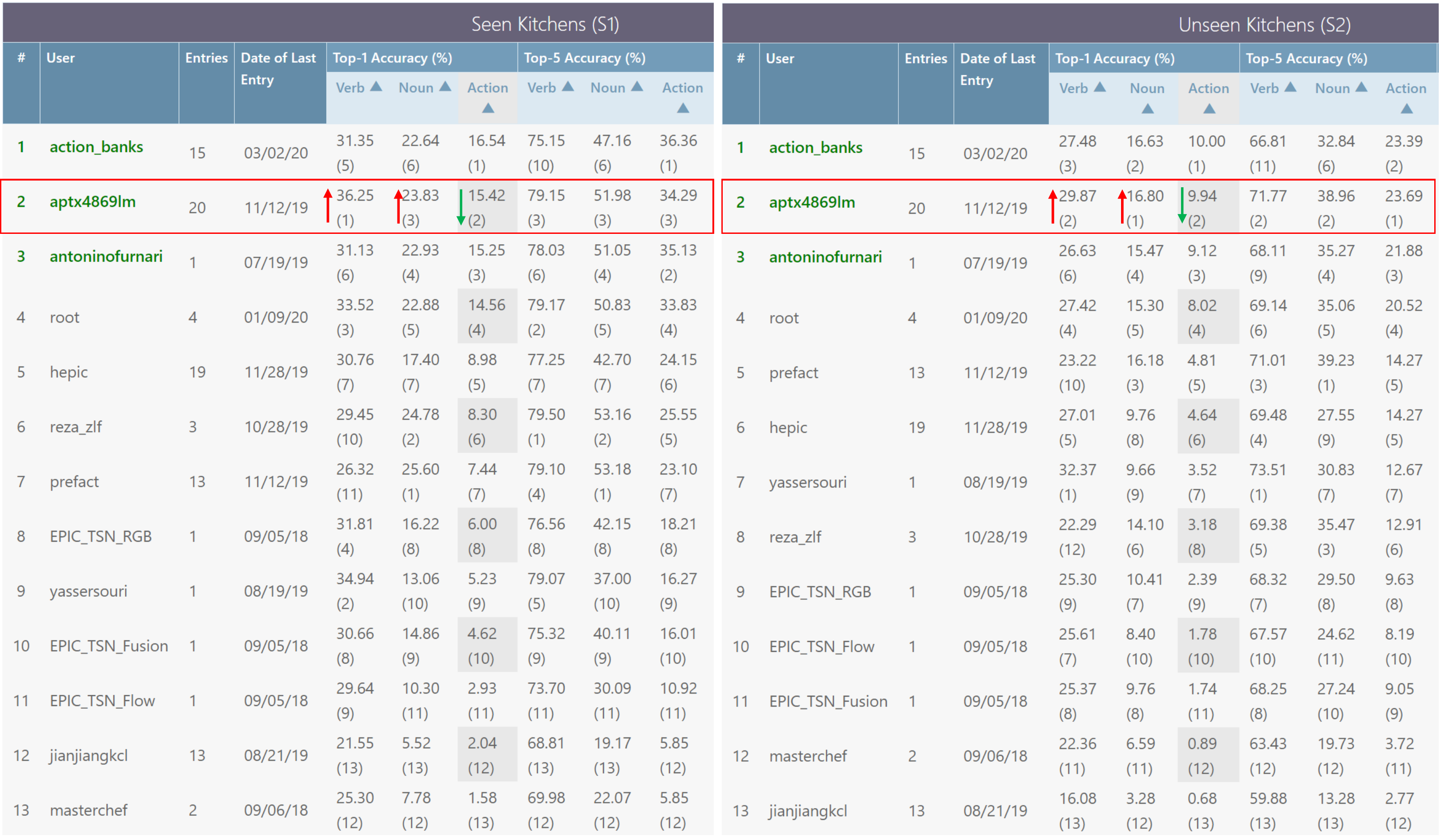}
\caption{Screenshot from Epic-Kitchens Anticipation Challenge. The user name of our proposed method is ``aptx4869lm''. The current rank1 team ``action\_banks'' is unpublished work, and lags behind of our method for both verb and noun prediction on both sets.  Note that user ``antonionfurnari'' refers to RULSTM in our main submission. They further improved the results reported in their paper.}
\label{fig:epic-results}
\end{figure*}

\section{Comparison of Experiment Setup to RULSTM}
\label{sec:s4}
\begin{table*} [t]
\caption{Comparison between our methods and previous state-of-the-art results RULSTM. See Sec.4.2 of our submission for discussion of Ous+Obj.}
\setlength{\tabcolsep}{2pt} 
\renewcommand{\arraystretch}{1} % Default value: 1
\centering
\scriptsize
\begin{tabular}{ccccc}
\multicolumn{1}{c}{{\textbf{Method}}}        &\multicolumn{1}{c}{{\textbf{Tasks}}} & \multicolumn{1}{c}{{\textbf{Training Supervision}}}   & \multicolumn{1}{c}{{\textbf{Testing Inputs}}} & \multicolumn{1}{c}{\textbf{End-to-End}}\\ \hline 
RULSTM [15]    & \makecell{Action Anticipation} & \makecell{Action Labels \\ Object Cls \& Boxes}   &  \makecell{RGB + Object Feat. \\ + Flow} & No
\\ \hline
Ours &\makecell{Action Anticipation\\ Visual Affordance \\ Motor Attention Pred} & \makecell{Action Labels \\ Hand \& Hotspots} & RGB & Yes\\ \hline
Ours+Obj &\makecell{Action Anticipation\\ Visual Affordance \\ Motor Attention Pred} & \makecell{Action Labels \\ Object Cls \& Boxes \\Hand \& Hotspots} & \makecell{RGB + Object Feat.} & No \\ 
\end{tabular}
\label{table:expsetup}
\end{table*}

We highlight our comparison to RULSTM. In Sec.4.2 of our submission, we contrast our method with previous state-of-the-art result RULSTM~\cite{furnari2019rulstm}. Here, we draw a more clear boundary between our method and RULSTM. In Table~\ref{table:expsetup}, we present the experiment setup of our method and RULSTM. Both RULTM and our model (Ours) use various supervisory signals for training, yet our model only needs RGB frames for inference and is end-to-end trainable. Ours+Obj model does require more training signals in comparison to RULSTM, yet it does not need optical flow for two-stream architecture. We have to point out that, from practical prospective, we care more about the data modality during testing time. Therefore, using more supervisory signals for training does not compromise the contribution of our method. Moreover, our method also address the challenging problem of motor attention prediction and interaction hotspots estimation.

\section{Epic-Kitchens Challenge Leaderboard}
\label{sec:s5}
Fig~\ref{fig:epic-results} presents a screenshot of the leaderboard from the EPIC-Kitchens Action Anticipation Challenge (\url{https://epic-kitchens.github.io/}).\footnote{Retrieved at March 13th, 2020.} The screenshot was acquired on the last day of supplementary material deadline. To date, our proposed method outperforms all published results by a large margin. Several very recent unpublished work (user id: ``action\_banks'', ``reza\_zlf'', ``hepic'', ``prefact'', ``root'' in Fig.~\ref{fig:epic-results}) also attempted at the EPIC-Kitchens Action Anticipation Challenge. The only work that outperforms our method is ``action\_banks''. Although ``action\_banks'' slightly outperforms our method for action prediction,  their results are worse than our method in terms of the verb and noun prediction.

%The only work that comes close to our method is ``action\_banks''. Their method works slightly better than ours on seen test set s1 action anticipation, while our method our outperforms them on seen test set s1 verb and noun anticipation, and all anticipation tasks on unseen test set s2.

\section{Experiment on Gaze Fixation Model}
\label{sec:s6}
\begin{table}[t]
\caption{Additional results on fixation based model. We contrast Gaze Only Model with baseline I3D model and our full model. }
\centering
{
\tablestyle{2pt}{1.0}
\setlength{\tabcolsep}{3pt} 
\renewcommand{\arraystretch}{1.05} 
\begin{tabular}{c|ccc}
Methods                                                         & Verb    & Noun  & Action \\ \hline

I3D-Res50   & 48.01/31.25 &	42.11/30.01 &34.82/23.20\\

Gaze Only$^\dagger$                            & 47.88/31.79	&43.83/33.42	&35.31/24.51 \\
Ours$^\dagger$                                                      &48.96/32.48	&45.50/32.73	&36.60/25.30

\end{tabular}
}
\label{table:ablation:gaze}
\end{table}

In this section, we present additional results on using gaze as attention distribution for visual anticipation. We follow [31] to replace motor attention and hotspots modules with a gaze module. We denote the resulting model as Gaze Only model. The experiments are conducted on EGTEA dataset, as gaze is not available on EPIC-Kitchens dataset. Gaze Only model improves the I3D-Res50 baseline by a notable margin. However, it lags behind our full model. This is because our model explicitly reasons about the future representation by making motor attention a first class player.

\section{Additional Qualitative Results}
\label{sec:s7}
Finally, we provide additional qualitative results. We included a video demo of our results as part of our supplementary materials. In this document, we also present more samples of predicted motor attention, interaction hotspots, and action labels in Fig~\ref{fig:vis-more}. The figure follows the same format as Fig.~3 in the submission. These results further show that our proposed motor attention module has the remarkable ability of ``imagining'' possible hand movements even without the presence of hands in the observed video segments. 
Another interesting observation is that the predicted distribution of interaction hostpots can be sparse in certain circumstances (e.g., ``Open Fridge'' or ``Take Condiment''). This is because of the stochastic patterns of human-object interaction: There might be multiple valid contact regions for interaction, especially when the future active object has a relatively large scale. This again shows the necessity of the stochastic units in our proposed method. 

As discussed in our submission, the occlusion and absence of active objects might make the anticipation problem extremely challenging even for humans. The failure cases in Fig.~\ref{fig:vis-more} also suggest that the anticipation model can be biased by on-going action. This is because current FPV datasets (especially EPIC-Kitchens) segment a continuous action into several same atomic actions to ensure all action segments have similar temporal dimension. For instance, A video clip of ``cutting onions'' for 20 seconds is segmented into 7 or 8 shorter clips all having the same ``cutting onions'' label. This increases the transition probability of staying in current action state, and thereby biases the model. Therefore, the ability of predicting when exactly the action will end is important for more accurate action prediction model. This task is also related to the action localization problem in the literature~\cite{gu2018ava}.

\begin{table*}[t]
\small
\def\arraystretch{1.3}
\setlength{\tabcolsep}{0.7pt}
\centering
\scalebox{0.71}{

\begin{tabular}{c|c|c|c|c|c|c}
\hline 
\multirow{2}{*}{\textbf{ID}} & \multirow{2}{*}{\textbf{Branch}}                                                       & \multirow{2}{*}{\textbf{Type}}                                                & \multirow{2}{*}{\begin{tabular}[c]{@{}c@{}}\textbf{Kernel Size}\\ THW,(C)\end{tabular}} & \multirow{2}{*}{\begin{tabular}[c]{@{}c@{}}\textbf{Stride}\\ THW\end{tabular}} & \multirow{2}{*}{\begin{tabular}[c]{@{}c@{}}\textbf{Output Size}\\ THWC\end{tabular}} & \multirow{2}{*}{\textbf{Comments (Loss)}}                                                  \\
&   &   &   &   &   &                                                                                    \\ \hline 
1   & \multirow{24}{*}{\begin{tabular}[c]{@{}c@{}}Backbone\\ (shared)\end{tabular}} 
& Conv3D    & 5x7x7,64     & 2x2x2     & 16x112x112x64 \\ \cline{3-7}
2  & & MaxPool1  & 2x3x3  & 2x2x2  & 8x56x56x64  &   \\ \cline{3-7}
3   &    & \begin{tabular}[c]{@{}c@{}}Layer1\\Bottleneck 0-2 \end{tabular}    & \begin{tabular}[c]{@{}c@{}}3x1x1,64\\ 1x3x3,64\\1x1x1,256\end{tabular} (3 times)     & \begin{tabular}[c]{@{}c@{}}1x1x1\\ 1x1x1\\1x1x1\end{tabular} (3 times)     & 8x56x56x256  &  \\ \cline{3-7}
4    & & MaxPool2  & 2x1x1  & 2x1x1  & 4x56x56x256  & \begin{tabular}[c]{@{}c@{}}Addition Pooling\\ Reduce Memory Usage\end{tabular} \\ \cline{3-7}
5   &    & \begin{tabular}[c]{@{}c@{}}Layer2\\Bottleneck 0 \end{tabular}    & \begin{tabular}[c]{@{}c@{}}3x1x1,128\\ 1x3x3,128\\1x1x1,512\end{tabular}     & \begin{tabular}[c]{@{}c@{}}1x1x1\\ 1x2x2\\1x1x1\end{tabular}     &   &  \\ \cline{3-7}
6   &    & \begin{tabular}[c]{@{}c@{}}Layer2\\Bottleneck 1-3 \end{tabular}    & \begin{tabular}[c]{@{}c@{}}3x1x1,128\\ 1x3x3,128\\1x1x1,512\end{tabular}  (3 times)    & \begin{tabular}[c]{@{}c@{}}1x1x1\\ 1x2x2\\1x1x1\end{tabular} (3 times) & 4x28x28x512 &  \\ \cline{3-7}
7   &    & \begin{tabular}[c]{@{}c@{}}Layer3\\Bottleneck 0 \end{tabular}   & \begin{tabular}[c]{@{}c@{}}3x1x1,256\\ 1x3x3,256\\1x1x1,1024\end{tabular}     & \begin{tabular}[c]{@{}c@{}}1x1x1\\ 1x2x2\\1x1x1\end{tabular}   &  &  \\ \cline{3-7}
8   &    & \begin{tabular}[c]{@{}c@{}}Layer3\\Bottleneck 1-5 \end{tabular}   & \begin{tabular}[c]{@{}c@{}}3x1x1,256\\ 1x3x3,256\\1x1x1,1024\end{tabular}  (5 times)     & \begin{tabular}[c]{@{}c@{}}1x1x1\\ 1x1x1\\1x1x1\end{tabular}   (5 times)  &4x14x14x1024  &  \\ \cline{3-7}

9   &    & \begin{tabular}[c]{@{}c@{}}Layer4\\Bottleneck 0 \end{tabular}    & \begin{tabular}[c]{@{}c@{}}3x1x1,128\\ 1x3x3,128\\1x1x1,512\end{tabular}     & \begin{tabular}[c]{@{}c@{}}1x1x1\\ 1x2x2\\1x1x1\end{tabular}     &  &  \\ \cline{3-7}
10   &    & \begin{tabular}[c]{@{}c@{}}Layer4\\Bottleneck 1-2 \end{tabular}    & \begin{tabular}[c]{@{}c@{}}3x1x1,128\\ 1x3x3,128\\1x1x1,512\end{tabular}  (2 times)    & \begin{tabular}[c]{@{}c@{}}1x1x1\\ 1x2x2\\1x1x1\end{tabular} (2 times)     &4x7x7x2048  &  \\ \cline{3-7} \hline

11 & \multirow{8}{*}{\begin{tabular}[c]{@{}c@{}}Motor \\Attention \\ Module\end{tabular}}       & \begin{tabular}[c]{@{}c@{}}Conv3d 1\\ (on Layer 2 feature)\end{tabular} & 1x3x3,128    & 1x1x1    & 4x28x28x128 &    \\ \cline{3-7}
12 &      & Conv3d 2   & 1x3x3,1 & 1x1x1  & 4x28x28x1 & \begin{tabular}[c]{@{}c@{}}KLD Loss\end{tabular}   \\ \cline{3-7}
13 &      & Maxpool 1  & 1x2x2 &1x2x2 &  4x14x14x1 & Guiding Interaction Hotspots \\ \cline{3-7} 
14                                      &                                                                               & \begin{tabular}[c]{@{}c@{}}Gumbel Softmax 1\\ (Sampling)\end{tabular}                                                        &                                                                              &                                                                         & 4x14x14x1                                                                              &      Sampling Motor Attention                                                                            \\ \cline{3-7} 
15 &      & Maxpool 2  & 1x4x4 &1x4x4 &  4x7x7x1 &  Guiding Action Anticipation \\ \cline{3-7} 

16                                      &                                                                               & \begin{tabular}[c]{@{}c@{}}Gumbel Softmax 2\\ (Sampling)\end{tabular}                                                        &                                                                              &                                                                         & 4x7x7x1                                                                              &      Sampling Motor Attention                                                                            \\ \hline
17 & \multirow{5}{*}{\begin{tabular}[c]{@{}c@{}}Interaction \\Hotspots \\ Module\end{tabular}}       & \begin{tabular}[c]{@{}c@{}}Weighted Pooling\end{tabular} &  & & 4x14x14x256 & \begin{tabular}[c]{@{}c@{}} With Sampled Motor Attention\end{tabular}    \\ \cline{3-7}

18 &  & \begin{tabular}[c]{@{}c@{}}Conv3d 1\\ (on Layer 3 Feature)\end{tabular} & 1x3x3,256    & 1x1x1    & 4x14x14x256 &    \\ \cline{3-7}

19 &      & Conv3d 2   & 1x3x3,1 & 1x1x1  & 4x14x14x1 & \begin{tabular}[c]{@{}c@{}}KLD Loss\end{tabular}   \\ \cline{3-7}
20 &      & Maxpool 1  & 1x2x2 &1x2x2 &  4x7x7x1 & Guiding Action Anticipation \\ \cline{3-7} 
21                                      &                                                                               & \begin{tabular}[c]{@{}c@{}}Gumbel Softmax \\ (Sampling)\end{tabular}                                                        &                                                                              &                                                                         & 4x7x7x1                                                                              &      Sampling Interaction Hotspots         \\ \hline

22 & \multirow{4}{*}{\begin{tabular}[c]{@{}c@{}}Action\\Anticipation \\ Module\end{tabular}}       & \begin{tabular}[c]{@{}c@{}}Weighted\\ Avg Pool\\ (on Final Feature)\end{tabular} & 4x7x7    & 4x7x7    & 1x1x1x1024 &   \begin{tabular}[c]{@{}c@{}} With Sampled Motor Attention\\ and Interaction Hotspots\end{tabular}  \\ \cline{3-7}
23                                      &                                                                               & Fully Connected                                                      &                                                                              &                                                                         & 1x1x1xN                                                                        &                                                                                  \\ \cline{3-7} 
24                                      &                                                                               & Softmax                                                              &                                                                              &                                                                         & 1x1x1xN                                                                & \begin{tabular}[c]{@{}c@{}}Cross Entropy Loss\\ (Action Anticipation)\end{tabular} \\ \hline

\end{tabular}}
\caption{Network architecture of our proposed model. We omit the residual connection in backbone ResNet-50 for simplification. }
\label{table:structure}
\end{table*}

\end{document}